%% file: root.tex
\documentclass[conference]{IEEEtran}
\IEEEoverridecommandlockouts
\usepackage{cite}
\usepackage{amsmath,amssymb,amsfonts}
\usepackage{algorithmic}
\usepackage{graphicx}
\usepackage{soul}
\usepackage{textcomp}
\usepackage{xcolor}
\usepackage{multirow}
\usepackage{makecell}
\def\BibTeX{{\rm B\kern-.05em{\sc i\kern-.025em b}\kern-.08em
    T\kern-.1667em\lower.7ex\hbox{E}\kern-.125emX}}
\usepackage{listings}
\usepackage{tcolorbox}
\usepackage{hyperref} 
\usepackage[normalem]{ulem}

\definecolor{promptbg}{RGB}{200,200,200}
\definecolor{responsebg}{RGB}{215,230,250}

\begin{document}

\IEEEaftertitletext{\vspace{-2.5\baselineskip}}

\title{TB or Not TB: Coverage-Driven Direct Preference Optimization for Verilog Stimulus Generation \\ 
}

\author{\IEEEauthorblockN{1\textsuperscript{st} Bardia Nadimi}
\IEEEauthorblockA{\textit{BCAICC} \\
\textit{U. of South Florida}\\
bnadimi@usf.edu}
\and
\IEEEauthorblockN{2\textsuperscript{nd} Khashayar Filom}
\IEEEauthorblockA{\textit{Core AI Department} \\
\textit{Cognichip Inc}\\
khashayar@cognichip.ai}
\and
\IEEEauthorblockN{3\textsuperscript{rd} Deming Chen}
\IEEEauthorblockA{\textit{Dept. of ECE} \\
\textit{U. of Illinois Urbana-Champaign}\\
dchen@illinois.edu}
\and
\IEEEauthorblockN{4\textsuperscript{th} Hao Zheng}
\IEEEauthorblockA{\textit{BCAICC} \\
\textit{U. of South Florida}\\
haozheng@usf.edu}
}

\maketitle

\begin{abstract}

With the rapid advancement of Large Language Models (LLMs), there is growing interest in applying them to hardware design and verification. 
Among these stages, design verification remains the most time-consuming and resource-intensive phase, where generating effective stimuli for the design under test (DUT) is both critical and labor-intensive.
We present {\it TB or not TB}, a framework for automated stimulus generation using LLMs fine-tuned through Coverage-Driven Direct Preference Optimization (CD-DPO). 
To enable preference-based training, we introduce PairaNet, a dataset derived from PyraNet that pairs high- and low-quality testbenches labeled using simulation-derived coverage metrics. 
The proposed CD-DPO method integrates quantitative coverage feedback directly into the optimization objective, guiding the model toward generating stimuli that maximize verification coverage.
Experiments on the CVDP CID12 benchmark show that {\it TB or not TB} outperforms both open-source and commercial baselines, achieving up to 77.27\% improvement in code coverage, demonstrating the effectiveness of Coverage-driven preference optimization for LLM-based hardware verification.

\end{abstract}




\input{sections/1-introduction}
\input{sections/3-method}
\input{sections/4-comparison_and_evaluation}
\input{sections/5-conclusion}
\input{sections/6-Acknowledgement}

\clearpage

\input{sections/References}
\end{document}

%% file: sections/1-introduction.tex
\vspace*{-8pt}
\section{Introduction and Motivation}
\vspace*{-4pt}
Since the introduction of the Transformer architecture~\cite{attentionIsAllYouNeed}, attention-based models have become central to modern AI. 
Models such as GPT~\cite{radford2018GPT}, BERT~\cite{BERT}, and LaMDA~\cite{LaMDA} have demonstrated the scalability and versatility of transformers across a wide range of language tasks, solidifying their role as the foundation of contemporary AI systems.

Building on these advances, the hardware design and verification community has increasingly explored Large Language Models (LLMs) to automate traditionally manual tasks in Electronic Design Automation (EDA). 
While early studies explored RTL code generation \cite{MEV-LLM, verithoughts, verimind, verirl} and design documentation \cite{docGen, mwscas}, recent work has shifted to design verification \cite{verilogCoder, AutoBench, CorrectBench, LLM4DV, llmaidedTbGen, assertionForge}, which remains one of the most critical and resource-intensive stages of the  design cycle (often more than half of the project effort). 

A core bottleneck in verification is generating high-quality stimuli within testbenches to exercise the design under test (DUT) across functional and corner-case scenarios. 
Manual testbench development is labor-intensive, error-prone, and requires significant expertise to balance coverage goals while avoiding over-constrained tests. 
Although Automatic Test Pattern Generation (ATPG) automates test pattern generation, it primarily targets manufacturing faults and relies on specific fault models. 
These challenges have motivated growing interest in LLM-driven verification to accelerate and scale testbench generation while preserving verification quality.

Recent research has increasingly explored the use of LLMs to automate hardware design verification, particularly testbench and stimulus generation. 
LLM4DV \cite{LLM4DV} introduced one of the first open-source frameworks for LLM-driven stimulus generation, showing that prompt engineering can achieve coverage competitive with constrained-random testing. 
AutoBench \cite{AutoBench} extended this direction with a fully automated pipeline that generates and evaluates testbenches from design descriptions using structural and self-checking mechanisms. 
CorrectBench \cite{CorrectBench} further added self-validation and self-correction capabilities, enabling iterative refinement and improving pass rates and coverage. 
VerilogReader \cite{VerilogReader} complemented these efforts by integrating LLMs into a coverage-directed loop that analyzes Verilog code and coverage reports to generate stimuli targeting uncovered behavior.

Collectively, these works show a progression in LLM-based hardware verification—from prompt-based stimulus generation (LLM4DV) to automated evaluation (AutoBench), self-correcting refinement (CorrectBench), and coverage-driven feedback integration (VerilogReader). 
Despite this progress, key limitations remain in robustness, generalization, and the systematic use of quantitative coverage signals during training. 
These gaps motivate the need for a principled, coverage-driven optimization framework that directly aligns model behavior with verification objectives, as pursued in this work.

By combining prompt engineering, fine-tuning, and coverage evaluation via EDA tools, LLMs can be adapted to generate high-quality verification stimuli. 
When trained with curated datasets and preference-based optimization, these models learn to produce testbenches that meaningfully exercise the DUT while reducing manual effort. 
This paradigm lowers verification overhead and improves coverage, enabling more scalable and reliable AI-driven verification flows.

In this work, we propose a preference-based training framework for automated stimulus generation using LLMs, where the model learns from relative preferences between pairs of outputs rather than absolute rewards, enabling it to identify testbenches that better satisfy verification objectives. 
Unlike reinforcement learning approaches that rely on continual simulator interaction, our method reformulates the task as supervised preference learning: candidate testbenches are generated and simulated offline, coverage metrics are collected via professional EDA tools, and the model is fine-tuned to distinguish high-coverage from low-coverage samples, thereby learning coverage-driven generation behavior from offline data.

This design offers several advantages. 
First, training is fully GPU-bound and simulator-free (during training), enabling stable and sample-efficient optimization through dense preference supervision. 
Second, the infrastructure remains lightweight by eliminating online simulation, while decoupling data generation from model optimization allows large-scale parallel simulation for coverage extraction and dataset expansion with minimal overhead. 
Finally, unlike online RL methods, the preference signals originate from a higher-quality commercial model, providing richer domain knowledge and enabling more effective alignment of the target model with verification objectives.

The proposed framework provides a practical and scalable foundation for LLM-driven verification, achieving a balanced trade-off between efficiency, stability, and coverage fidelity. 
Our results show that preference-based optimization with offline coverage feedback is an effective step toward autonomous testbench generation. 
This paper makes the following key contributions:
\vspace*{-4pt}

\begin{itemize}
    \item [1)] We introduce the first framework for automated SystemVerilog stimulus generation that replaces simulator-in-the-loop reinforcement learning with offline supervised preference learning. By simulating testbenches once and storing their coverage metrics, the approach enables fully GPU-bound, scalable, and simulation-free training.
    \item [2)] We developed a coverage-driven DPO (CD-DPO) framework based on standard DPO~\cite{rafailov2023direct} that incorporates quantitative feedback from code, branch, and functional coverage metrics. This guides the model toward generating high-coverage, robust stimuli and links preference optimization with coverage-driven verification goals.
    
    \item [3)] We introduce PairaNet\footnote{\url{https://huggingface.co/datasets/bnadimi/TB_or_Not_TB}}, the first open-source dataset for coverage-driven DPO training. Derived from PyraNet, it provides paired testbenches labeled with simulation-based coverage metrics, enabling preference learning from quantitative verification feedback. Its public release supports reproducible, data-driven research in LLM-based hardware verification.
    
    
    
    \item [4)] Experimental results show that the proposed {\it TB or not TB} model achieves up to 77.27\% improvement in code coverage compared to the vanilla Qwen3 model and 56.78\% improvement compared to Claude Sonnet 3.5 when evaluated on the best result among 20 generations. 
\end{itemize}

The remainder of this paper is organized as follows. 
Section II presents our proposed methodology in detail. 
Section III reports the experimental results along with ablation studies and finally, Section IV concludes the paper.

%% file: sections/3-method.tex
\vspace*{-2pt}
\section{Methodology and Dataset}
\vspace*{-2pt}

This section outlines the proposed stimulus generation framework, reviews the DPO foundations it builds upon, describes the construction of the preference-based dataset, and details the CD-DPO technique incorporating quantitative coverage feedback for improved testbench generation.

\subsection{Overview of the Framework}
\begin{figure*}[t]
    \centering 
    \includegraphics[width=0.95\textwidth]{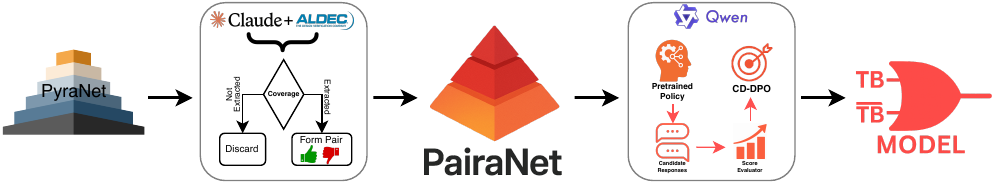}
    \caption{Overall {\it TB or not TB} Architecture.}
    \label{fig:overallTBnotTBArchitecture}
    \vspace*{-14pt}
\end{figure*}

\vspace*{-4pt}
The proposed framework follows a structured pipeline designed for preference-based optimization of testbench generation. 
As illustrated in Fig.~\ref{fig:overallTBnotTBArchitecture}, the process begins with the PyraNet dataset \cite{PyraNet}, which provides the foundational design samples used for dataset curation. 
We selected PyraNet as the base dataset because it provides a large, diverse collection of hardware designs with structured specifications and quality annotations, making it well-suited for constructing preference-labeled training data.
From this base, we construct a preference-ready dataset named PairaNet, containing paired examples of high- and low-quality testbenches labeled according to coverage-driven criteria.
Subsequently, the PairaNet dataset is employed to fine-tune an open-source Qwen language model, enabling it to internalize coverage-aware stimulus generation behavior. 
This fine-tuning process aligns the model’s generation preferences toward producing verification stimuli that yield higher coverage.
The resulting fine-tuned model, referred to as {\it TB or not TB}, embodies the learned coverage-driven verification knowledge. 
To evaluate its performance, the model is benchmarked using Section 12 (CID12) from NVIDIA’s CVDP dataset \cite{CVDP}, a standard evaluation suite for design verification tasks.
Each component of this framework including preference optimization fundamentals, dataset curation, and the proposed preference training is discussed in detail in the following subsections.

\subsection{DPO Fundamental Concepts}
\vspace*{-4pt}
LLMs have demonstrated remarkable capabilities in reasoning, code generation, and adaptation to human preferences through alignment techniques such as Reinforcement Learning from Human Feedback (RLHF) \cite{RLHF}. 
However, RLHF often suffers from high computational cost and instability due to its reliance on reinforcement learning components. 
To address these limitations, recent research has introduced preference-based optimization methods such as DPO, which provides a more direct and stable alternative to RLHF.

DPO formulates preference learning as a supervised fine-tuning problem guided by pairwise comparisons between model outputs \cite{rafailov2023direct}. 
Given a dataset of preferred and non-preferred samples $(x,y_w,y_l)$, DPO optimizes the model parameters $\theta$ to increase the likelihood of the preferred response $y_w$ while decreasing that of the non-preferred response $y_l$. 
The DPO loss is defined as
\vspace*{-12pt}

\begin{equation} 
\label{eqn1}
\begin{split}
    \mathcal{L}_{\rm {DPO}}(\pi_{\theta};\pi_{ref}) = & -\mathbb{E}_{(x,y_w,y_l)\sim\mathcal{D}}[\log \sigma (\beta r(y_w|x) \\
                                                & -\beta r(y_l|x))] \\
\end{split}
\end{equation}
where $r(y_w|x)= \log \frac{\pi_\theta (y_w|x)}{\pi_{\rm{ref}}(y_w|x)}$ is an implicit reward parameterization for the preferred sample, $r(y_l|x)= \log \frac{\pi_\theta (y_l|x)}{\pi_{\rm{ref}}(y_l|x)}$ is an implicit reward parameterization for the not-preferred sample, $\pi_\theta$ is the current model policy, $\pi_{\rm{ref}}$ is the reference model, $\sigma(.)$ is the sigmoid function, and $\beta$ controls the sharpness of preference scaling.
Unlike other RLHF methods, DPO eliminates the need for an explicit reward model and online policy gradient updates, leading to more stable and computationally efficient training. 
In the context of this work, DPO allows the model to directly learn which testbench samples produce higher coverage, effectively aligning the generator toward coverage-driven preferences.


Recent works have extended DPO into several interesting variants to address its limitations in handling diverse or noisy feedback signals. 
In \cite{multi-PO}, Multi-Preference Optimization (MPO) is introduced as well as various modifications to the loss function such as switching from sigmoid to another link function, forgoing the reference model, adding a length normalization etc. (see \cite{zhao2024rainbowpo} for a unified framework and a review of these variants.)

In summary, DPO provides a principled framework for aligning LLMs using structured and potentially multi-dimensional feedback. 
In this work, these preference-based optimization techniques form the foundation for training our testbench generator to favor verification stimuli that achieve higher coverage and exhibit functional robustness.

\vspace*{-4pt}

\subsection{Dataset Construction for Preference-Based Training}

To enable preference-based fine-tuning, we developed a multi-stage data curation pipeline that constructs paired examples of high- and low-quality testbenches. 
The process builds upon the PyraNet dataset \cite{PyraNet}, in which entries are categorized into six predefined quality tiers, ranging from Layer 1 (highest quality) to Layer 6 (lowest quality).
Each PyraNet sample includes design specifications, Verilog code, and associated metadata such as ranking scores. 
In the first stage, candidate testbenches were generated using Anthropic’s Claude Sonnet 3.5 model \cite{sonnet35} across Layers 1–5 of PyraNet; Layer 6 was excluded due to its un-compilable samples. 
For every design instance, two testbench generations were produced with distinct temperature settings to promote diversity in the generated stimuli.
Next, the generated candidates were processed through Verilator’s linting utility to ensure syntactic correctness. 
Pairs in which both generated testbenches were faulty were excluded from the data curation process.
Then, all compilable candidates were simulated using Aldec Riviera-Pro \cite{aldec}, from which three key coverage metrics were extracted: code, branch, and functional coverage.
We focus on code, branch, and functional coverage because together they capture complementary aspects of verification quality: structural execution, control-flow exploration, and user-defined functional intent. 
Using all three provides a more comprehensive and reliable assessment of how effectively generated stimuli exercise the design.
For each pair, we computed the average of these coverage metrics to establish a quantitative comparison. 
The testbench achieving the higher average coverage was labeled as the chosen sample, while the one with lower coverage (or that failed to compile) was labeled as the rejected sample.

The resulting dataset, termed PairaNet, extends the original PyraNet dataset \cite{PyraNet} by introducing preference pairs that differentiate high-coverage testbenches from low-coverage ones.
Each entry in PairaNet thus contains a triple of design prompt, preferred testbench, and rejected testbench (plus the metadata from PyraNet). 
For the preferred and rejected testbeches, the corresponding score is also provided.
This curated dataset serves as the foundation for CD-DPO training, allowing the model to internalize coverage-driven distinctions between high- and low-quality testbenches. 
In total we have 182,870 (pairs including a chosen and rejected testbenches) in the dataset.
Through this alignment process, the proposed framework progressively learns to generate verification environments that are both syntactically correct and coverage-efficient.
\vspace*{-6pt}

\subsection{Stimulus Generator Model Training}
Building upon the foundational principles of DPO, we propose a coverage-driven variant tailored for automated testbench generation. 
While standard DPO aligns the model with human or synthetic preferences through pairwise comparisons, our modified approach integrates coverage-based feedback from hardware simulation tools to guide learning toward high-quality verification outcomes.

In our framework, as mentioned before, each training sample consists of a hardware design prompt $x$ and two generated testbenches, a preferred sample $y_w$ and a non-preferred sample $y_l$. 
Instead of relying on subjective preference labels, we employ objective coverage metrics, including code, branch, and functional coverage, extracted from Aldec Riviera-Pro EDA tool. 
These metrics are averaged into a single coverage-driven preference score, which determines the relative ranking between the two samples. 
Testbenches that achieve higher coverage are marked as preferred.

To integrate this quantitative signal into the optimization process, we modify the conventional DPO loss to incorporate weighted preference scaling. 
This allows the training dynamics to place greater emphasis on samples with larger coverage differentials, effectively rewarding testbenches that contribute more significantly to verification completeness. 
The overall loss function can be expressed as:
\vspace*{-4pt}

\begin{equation} 
\label{eqn2}
\begin{split}
    \mathcal{L}_{{\rm CD\text{-}DPO}}(\pi_{\theta};\pi_{ref}) = & -\mathbb{E}_{(x,y_w,y_l)\sim\mathcal{D}}[\log \sigma (\beta^* r(y_w|x) \\
                                                & -\beta^* r(y_l|x))] \\
\end{split}
\end{equation}
where $\beta^* = \beta f(s_p - s_{np})$ is the replaced coefficient based on the coverage percentages, $s_p$ is the average coverage score for the preferred sample, $s_{np}$ is the average coverage score for the non-preferred sample, and $f(.)$ is an increasing function to normalize the difference of the coverage scores. 
Our modified DPO procedure supports per-sample weighting, allowing adaptive control of the $\beta$ parameter or gradient magnitude based on coverage quality. 
This design makes the optimization process more stable and data-efficient, especially when coverage values exhibit high variance across samples.
This formulation allows the model to balance syntactic correctness with verification effectiveness, because the preference signals reflect both the ability to compile and the achieved coverage, encouraging the model to favor outputs that satisfy structural requirements while also exercising more of the design’s behavior. As a result, the model is guided to generate testbenches that not only compile successfully but also attain higher functional coverage.

To see this, note that the contribution to the gradient of the modified loss~\eqref{eqn2} for a data point $(x, y_{w}, y_{l})$ is
\begin{equation*}
\begin{split}
-\beta^*\sigma (\beta^* r(y_l|x) -\beta^* r(y_w|x))
\big(\nabla\log\pi_{\theta}(y_w|x) \\- \nabla\log\pi_{\theta}(y_l|x)\big)
\end{split}
\end{equation*}
(cf. gradient expression in~\cite{rafailov2023direct}). Thus, a single SGD step adds a positive multiple of $\beta^*\sigma(\beta^* r(y_l|x) -\beta^* r(y_w|x))\left(\nabla\log\pi_{\theta}(y_w|x)-\nabla\log\pi_{\theta}(y_l|x)\right)$ to the parameters $\theta$.
Recall that $\beta^* = \beta f(s_p - s_{np})$ is the weight associated with this preference pair. A large score difference $s_{p} - s_{np}$ corresponds to an ``obvious'' pair and results in a large~$\beta^*$, whereas ambiguous pairs yield smaller~$\beta^{*}$.
Our modified DPO uses this mechanism to adjust the update strength: If the model currently disagrees with the preference based on its implicit reward, i.e.\ $r(y_{l}\mid x) - r(y_{w}\mid x) > 0$, then $\sigma(\beta^*(r(y_{l}\mid x)-r(y_{w}\mid x)))$ is relatively large and \emph{increases} with~$\beta^*$.
For clear (high-$\beta^{*}$) pairs, the gradient magnitude is therefore larger, causing $\theta$ to be updated more aggressively to correct an easy mistake.

Overall, the proposed coverage-Driven DPO approach provides a principled mechanism for aligning LLM-generated testbenches with quantitative verification objectives. 
By directly incorporating coverage feedback into the preference optimization loop, our model learns to generate testbenches that not only conform to design intent but also maximize verification completeness. 
Finally, we point out that there are variants of DPO with dynamic $\beta$ coefficient or with $\beta$ being multiplied by an instance-level factor in the literature, e.g. AlphaDPO \cite{wualphadpo}, $\beta$-DPO \cite{wu2024beta}, SimPO \cite{meng2024simpo},  WPO \cite{zhou2024wpo} and $\epsilon$-DPO \cite{lee2025kl}. 
Nevertheless, their loss functions are not quite the same as \eqref{eqn2}. 
Furthermore, to the best of our knowledge, applying a preference-based optimization algorithm to hardware verification is non-trivial and has not been explored in prior work. 
Verification signals derived from EDA tools and coverage metrics are discrete, sparse, and highly domain-specific, making their integration into a DPO-style framework challenging. 
In particular, incorporating a coverage-driven scaling of $\beta$ requires mapping simulator-derived quality differences into a stable preference-learning objective, making this formulation both technically demanding and novel within the hardware verification domain.

%% file: sections/4-comparison_and_evaluation.tex
\vspace*{-8pt}
\section{Evaluation and Discussion}
\vspace*{-6pt}


\subsection{Baseline Models}
\vspace*{-4pt}
We selected Anthropic’s Claude Sonnet-3.5 model \cite{sonnet35} for dataset curation due to its strong performance in structured code generation and instruction adherence. 
Compared to other available LLMs, Claude Sonnet-3.5 demonstrates superior consistency in producing syntactically correct and logically coherent SystemVerilog code, which is essential for large-scale testbench generation.
For the pre-trained open-source model, we selected the Qwen3 family, specifically the 4B, 8B, and 14B variants \cite{qwen3}. 
The primary reason for this choice is that the Qwen3 models demonstrate a strong balance between instruction-following capability, code generation proficiency, and computational efficiency, making them well-suited for hardware verification tasks. 
Furthermore, the availability of multiple model sizes also enables scaling studies that assess how model capacity affects coverage-driven preference alignment and testbench quality.
\vspace*{-6pt}

\subsection{Experiments}
\subsubsection{CVDP CID12 explanation}

\vspace*{-4pt}
To evaluate the effectiveness of the proposed method, we utilized the CID12 task from the CVDP benchmark \cite{CVDP}, a recent and comprehensive suite designed for assessing verification performance in hardware design automation.
It is important to note that the original CVDP framework utilizes Cadence Xcelium \cite{cadence} for simulation of tasks CID12, CID13, and CID14. 
However, due to limited access to Cadence tools, all experiments in this work were conducted using Aldec’s Riviera-Pro \cite{aldec}, which provides comparable coverage analysis capabilities.
While the original CVDP benchmark reports only code coverage for CID12, we extend this evaluation to include branch coverage, and functional coverage, providing a more comprehensive assessment of verification quality.
Additionally, the CVDP benchmark defines a binary pass/fail threshold for CID12 based on coverage levels. 
In contrast, our evaluation reports average coverage values across all coverage types, reflecting the continuous nature of coverage-driven improvement rather than discrete pass/fail outcomes.
As a result, the numerical results presented in this paper are not directly comparable to those in the original CVDP report, but they offer a more detailed and practical analysis of testbench quality under coverage-driven preference optimization. 

\vspace*{-2pt}

\subsubsection{Experiment setting}
In our experiments, the hyperparameter $\beta$ was set to $0.2$, and the function $f(\cdot)$ denotes a normalization operator that scales all scores to the range $[0, 1]$. 
Model training was conducted on eight NVIDIA H200 GPUs, requiring approximately 36 hours for the largest model configuration (Qwen3\_14B).
It should be mentioned that the results reported in Figures~\ref{fig:mean@1} to \ref{fig:ablation2} are based on the proprietary dataset that will be explained in Section \ref{ablationStudies} {\bf RQ4 answer}.

\subsubsection{Results Explained}

To better interpret and analyze the experimental outcomes, we formulated four research questions aimed at systematically evaluating different aspects of the proposed framework.  
{\bf RQ1:} How effectively does the proposed method improve verification coverage compared to baseline models?
{\bf RQ2:} How well does the framework scale across different model sizes in terms of efficiency?
{\bf RQ3:} How does the proposed coverage-driven DPO (CD-DPO) compare to standard DPO and Supervised Fine-Tuning (SFT) approaches in aligning model behavior with verification objectives?
{\bf RQ4:} How sensitive is the model’s performance to variations in training dataset quality?
Due to the analytical nature of Research Questions 3 and 4, their corresponding evaluations and discussions are presented in the \nameref{ablationStudies} (subsection C) for improved clarity and depth of analysis.

\begin{figure}[h]
    \vspace*{-6pt}
    \centering
        \includegraphics[width=0.95\columnwidth]{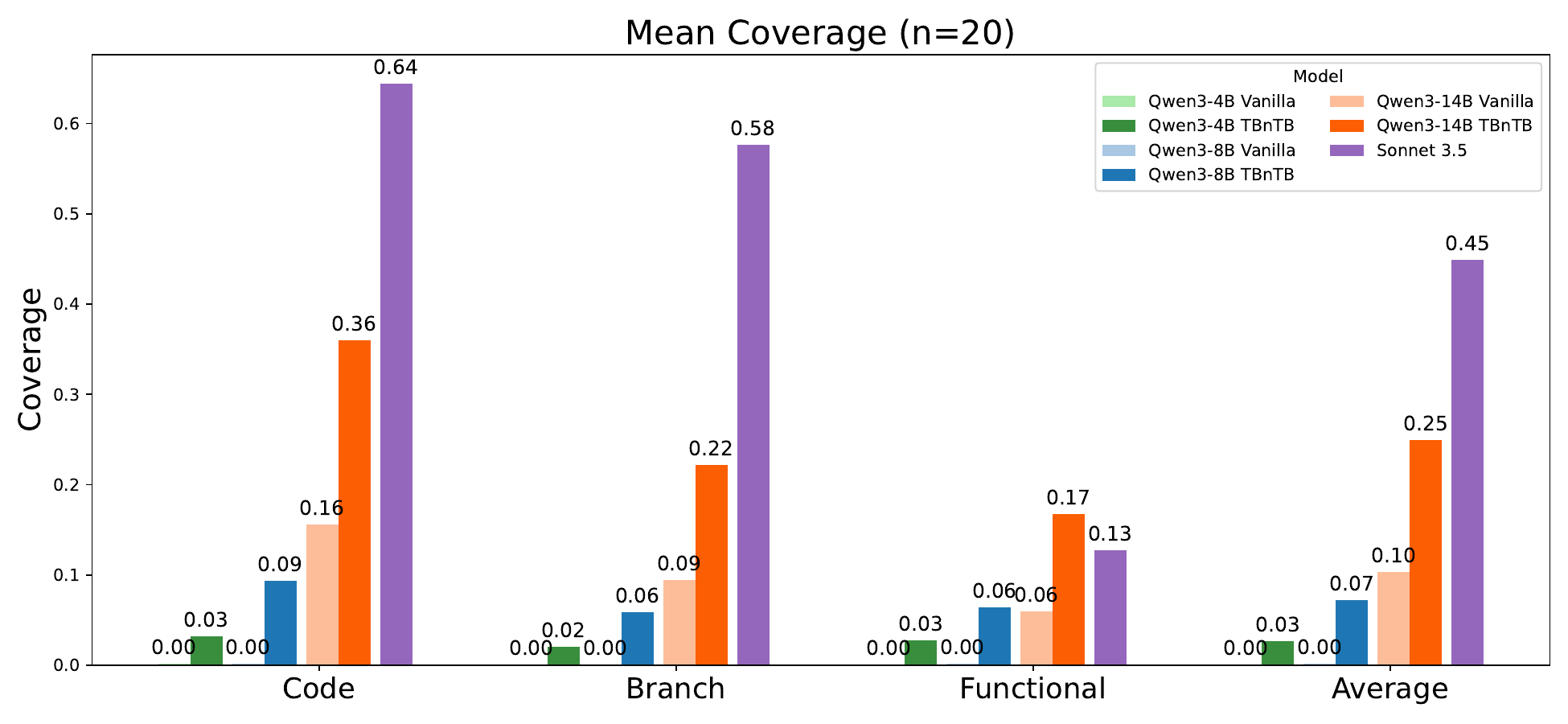}
    \vspace*{-12pt}
    \caption{Mean results for $n=20$ individual generations}
    \label{fig:mean@1}
    \vspace*{-12pt}
\end{figure}

\begin{figure}[h]
    \centering
        \includegraphics[width=0.95\columnwidth]{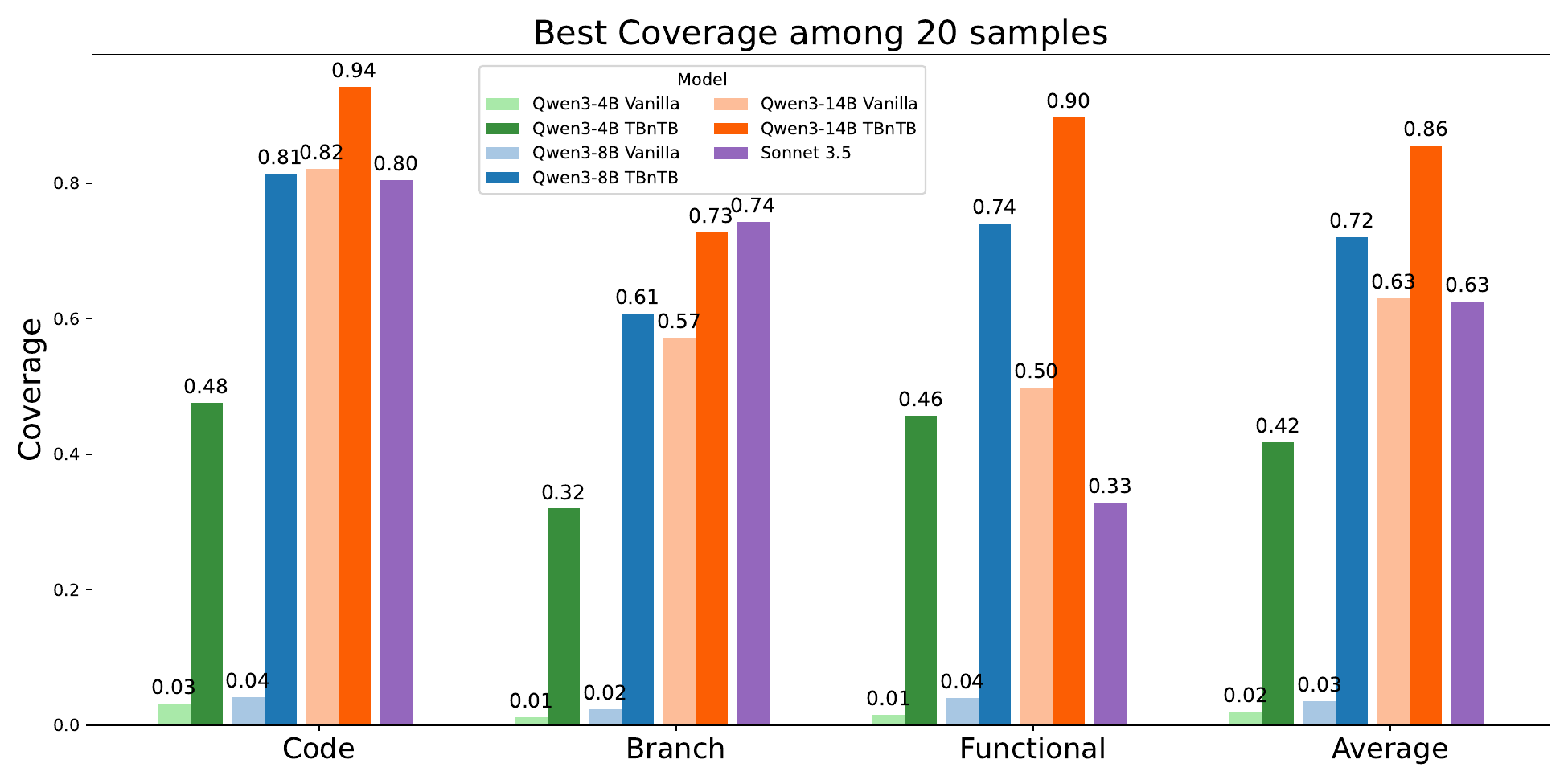}
    \vspace*{-10pt}
    \caption{Best results among $n=20$ individual generations}
    \label{fig:best@20}
    \vspace*{-8pt}
\end{figure}

{\bf RQ1 answer}:
To comprehensively assess the effectiveness of {\it TB or not TB}, we conducted a series of experiments based on CVDP CID12. 
In our experiments, each model was evaluated across 20 independent generations per design. 
The {\tt mean of 20} refers to the average coverage achieved across these 20 generations, reflecting the model’s overall consistency and reliability. 
The {\tt best of 20} denotes the highest coverage obtained among the 20 generations, capturing the model’s peak performance potential under optimal sampling conditions.
The results of these evaluations are summarized in Figures~\ref{fig:mean@1}–\ref{fig:best@20}.

We compared the performance of our model against multiple baselines, including the vanilla Qwen3 models (4B, 8B, and 14B variants) and Anthropic’s Claude Sonnet-3.5.
The results demonstrate significant improvements achieved by {\it TB or not TB} over both vanilla and commercial baselines.
Specifically, the proposed CD-DPO framework allows the model to achieve markedly higher coverage (up to a 77.27\% increase over Qwen3\_8B Vanilla) and reduce the performance gap between open-source and commercial models such as Sonnet-3.5.

It is particularly notable that this improvement is achieved despite the large parameter gap between Qwen3 and commercial models, highlighting the efficiency of our fine-tuning strategy.
Furthermore, when considering test-time compute (TTC) strategies \cite{ttc}, the results in Figure~\ref{fig:best@20} show that {\it TB or not TB} can achieve superior performance over commercial models when multiple generations are evaluated (up-to 56.78\% improvement in functional coverage when comparing Qwen3\_14B {\it TB or not TB} vs Sonnet 3.5). 
In other words, with lower computational cost and multiple testbench generations, our model can surpass the results of high-end proprietary models.
Although Anthropic has not disclosed the exact parameter count of Claude Sonnet 3.5, models of this scale are expected to be significantly larger than open-source counterparts such as the Qwen series. 
Consequently, inference with Sonnet 3.5 incurs substantially higher computational and financial costs.
These findings confirm the effectiveness of the proposed CD-DPO framework in improving both stimulus diversity and verification quality, demonstrating that high-quality verification models can be developed without relying on large-scale proprietary systems.

It is important to note that, in the context of functional coverage, both open-source and commercial models exhibit the capability to automatically generate covergroups, the SystemVerilog constructs used to define coverage points and bins that capture key functional behaviors of the design.
However, due to the inherently qualitative nature of evaluating coverage model quality, establishing a ground truth for direct comparison remains highly challenging. 
This difficulty is further reflected in the CVDP benchmark CID12, which also omits reference coverage models for this reason.
Consequently, the functional coverage results reported here should be interpreted solely as an indication of the models’ ability to produce syntactically valid and compilable covergroups, rather than as a measure of the semantic quality or completeness of the generated coverage models.

Although the mean performance across 20 runs of our model is lower than that of the commercial baseline (see Figure~\ref{fig:mean@1}), the best-performing instance among these runs surpasses the best result obtained from the commercial model. 
This observation suggests that while the commercial model exhibits greater consistency and stability across generations, our approach demonstrates a higher performance ceiling, indicating stronger potential when favorable sampling conditions (such as temperature settings or decoding configurations that lead to higher-quality outputs) or optimal generation trajectories (in which the model follows reasoning and synthesis paths aligned with the design’s coverage objectives) are achieved.
The variability in our model’s outputs may arise from factors such as sensitivity to initialization, or limited fine-tuning stability, highlighting an opportunity for future work to improve robustness while maintaining peak performance.
It is worth noting that, since Claude Sonnet 3.5 represents a substantially stronger model compared to current state-of-the-art open-source alternatives, we limit our comparison to this model and do not include results from other open-source baselines.
\vspace*{-6pt}

\vspace*{-12pt}

{\bf RQ2 answer:} 
As illustrated in Figures~\ref{fig:mean@1} and \ref{fig:best@20}, a comparison across the fine-tuned Qwen3 models with 4B, 8B, and 14B parameters demonstrates that the proposed approach scales effectively with model size. 
Larger models consistently achieve higher coverage and exhibit more stable performance across generations, indicating that the CD-DPO framework benefits from increased model capacity without compromising training stability or efficiency. 
This scaling trend also helps narrow the performance gap with much larger and stronger proprietary models such as Sonnet 3.5. 
As the model size increases, the combination of greater representational capacity and coverage-driven optimization allows Qwen-based models to better capture complex verification behaviors and respond more effectively to coverage feedback. 
Consequently, the CD-DPO framework provides a clear path toward matching or surpassing commercial-grade performance with potentially much less computational cost.
\vspace*{-8pt}

\subsection{Ablation Studies}
\label{ablationStudies}

\begin{figure}[h]
    \centering
        \includegraphics[width=0.95\columnwidth]{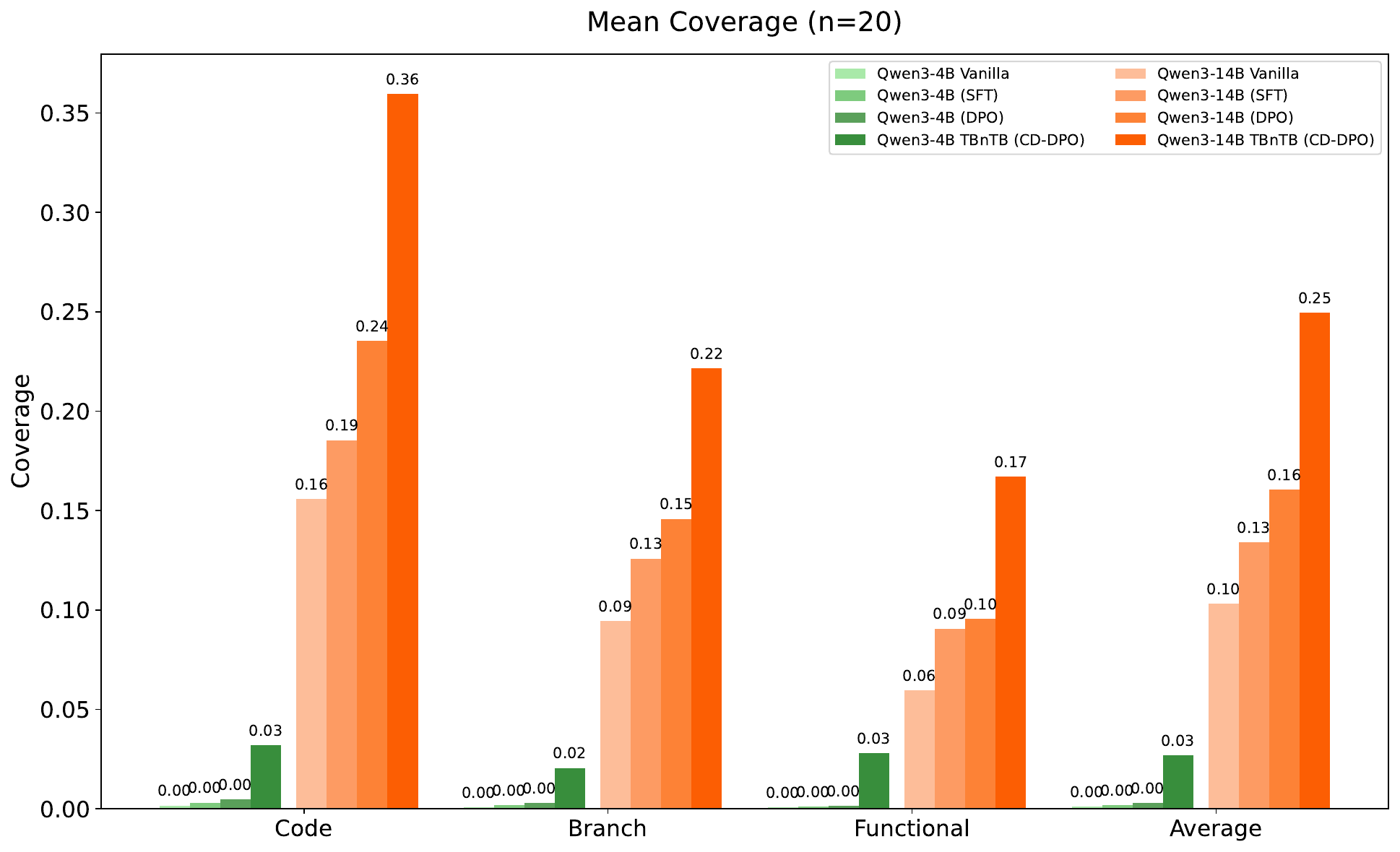}
    \vspace*{-12pt}
    \caption{Mean results for $n=20$ individual generations for Vanilla vs SFT vs standard DPO vs CD-DPO}
    \label{fig:ablation}
    \vspace*{-10pt}
\end{figure}

\begin{figure}[h]
    \centering
        \includegraphics[width=0.95\columnwidth]{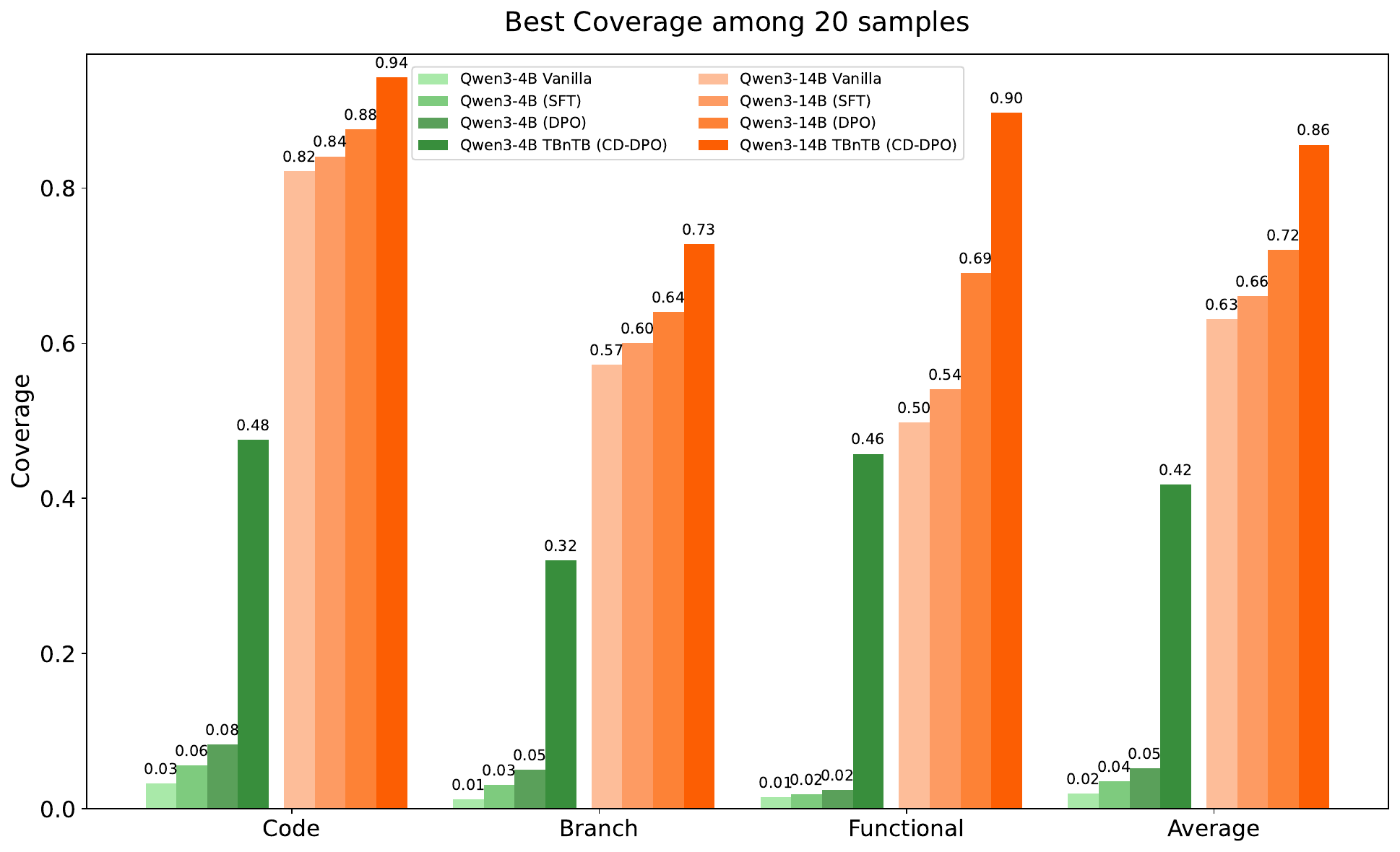}
    \vspace*{-10pt}
    \caption{Best results among $n=20$ individual generations for Vanilla vs SFT vs standard DPO vs CD-DPO}
    \label{fig:ablation2}
    \vspace*{-12pt}
\end{figure}

\begin{figure}[h!]
    \centering
        \includegraphics[width=0.95\columnwidth]{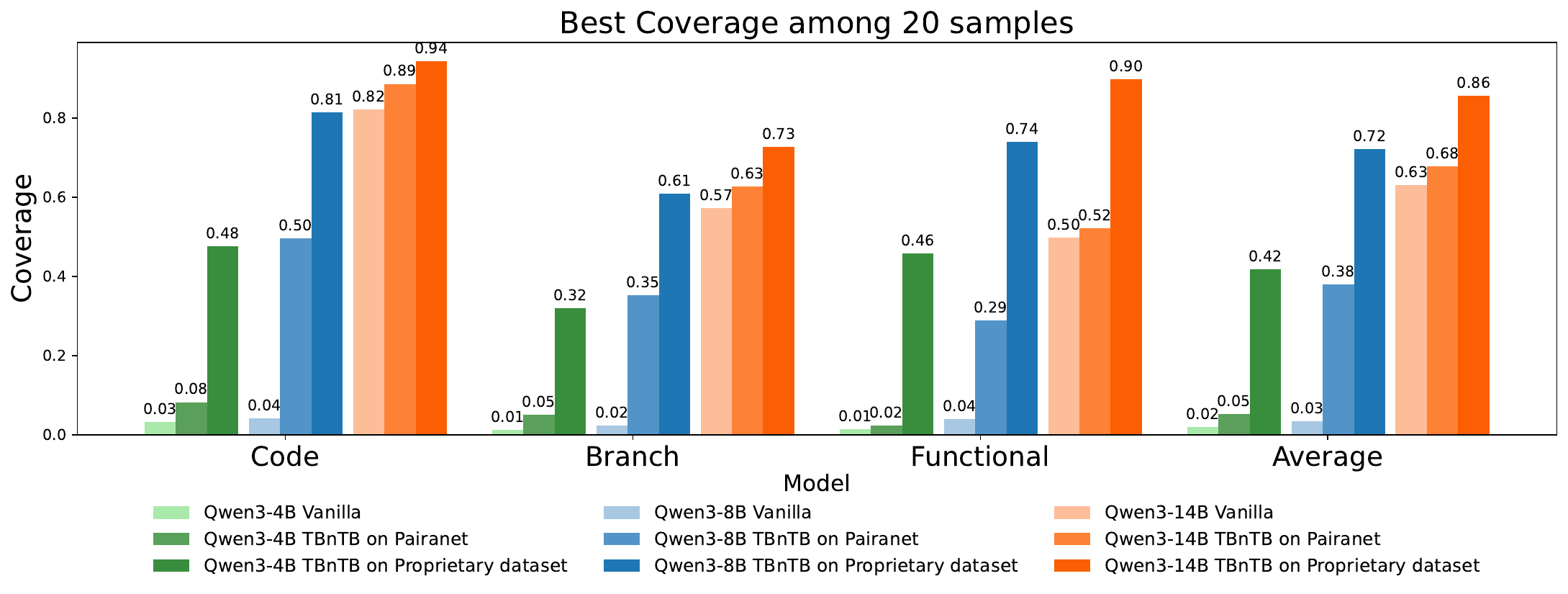}
    \vspace*{-10pt}
    \caption{PairaNet vs Proprietary dataset results}
    \label{fig:scaling}
    \vspace*{-12pt}
\end{figure}

\vspace*{-2pt}
To further validate the effectiveness of the proposed CD-DPO framework, we conducted a series of ablation experiments. 
{\bf RQ3 answer:}
We trained three additional models for comparison: (1) a model fine-tuned solely using Supervised Fine-Tuning (SFT) without any preference-based optimization, (2) a model trained with an alternative loss function, and (3) a model trained with the standard DPO loss function without our proposed coverage-driven modifications as in \eqref{eqn2}.
As illustrated in Figures~\ref{fig:ablation} and \ref{fig:ablation2}, the model trained with the CD-DPO method consistently outperforms all baselines across multiple evaluation metrics. 
The SFT-only model shows limited improvement, indicating that exposure to higher-coverage samples alone is insufficient for coverage-oriented learning. 
Similarly, the models trained with the unmodified loss formulations fail to reach the same coverage quality levels as the proposed approach.
These results clearly demonstrate the effectiveness of integrating coverage-driven feedback into the DPO framework, confirming that our proposed coverage-driven loss function enables the model to generate higher-quality and more verification-effective testbenches.
It should be noted that in Figures~\ref{fig:ablation} and \ref{fig:ablation2}, all experiments were additionally repeated using the Qwen3\_14B as well as Qwen3\_4B models. 
This was done to further analyze the effect of model scale, as the performance differences among the Vanilla, SFT, and DPO variants were not distinctly observable in the initial results using Qwen3\_4B model.
{\bf RQ4 answer:}
In addition to PairaNet, we curated a proprietary high-quality dataset containing samples with richer design specifications and significantly higher code, branch, and functional coverage. 
This dataset was used to investigate the impact of dataset quality on the model’s training dynamics and overall performance.
Figure~\ref{fig:scaling} illustrates the impact of training dataset quality on model performance. 
The training procedure for {\it TB or not TB} was kept identical across both the PairaNet and proprietary datasets, ensuring that any observed differences arise solely from variations in dataset quality. 
The results clearly demonstrate that models trained on the higher-quality proprietary dataset achieve superior coverage and more stable convergence behavior, highlighting the importance of dataset richness (particularly in terms of detailed specifications and coverage diversity) in enhancing the effectiveness of coverage-driven preference optimization.

%% file: sections/5-conclusion.tex
\vspace*{-8pt}
\section{Conclusion and Future Works}
\vspace*{-2pt}

In summary, {\it TB or not TB} is a coverage-driven LLM framework that extends DPO into a CD-DPO formulation to align testbench generation with quantitative coverage metrics. Trained on datasets specifically curated for preference-based optimization, it achieves higher coverage and efficiency than both vanilla and commercial models on the CVDP CID12 benchmark when evaluated under the {\tt best of 20} setting.

Finally for future work, we plan to extend this framework in several directions. 
We aim to incorporate multi-objective optimization to jointly improve coverage and assertion robustness, and to integrate formal verification signals alongside simulation-based metrics for richer supervision.


%% file: sections/6-Acknowledgement.tex
\vspace*{-4pt}
\section*{Acknowledgment}

\vspace*{-4pt}

This work was supported by the National Science Foundation under Grant No. 2434247.
Any opinions, findings, and conclusions or recommendations expressed in this material are those of the authors and do not necessarily reflect the views of the funding agency. 
Cognichip and Aldec provided computational resources and licenses critical for this research. 



\vspace*{-10pt}